\pdfoutput=1
\documentclass[11pt]{article}

\usepackage[]{ACL2023}

\usepackage{times}
\usepackage{latexsym}

\usepackage[T1]{fontenc}
\usepackage[utf8]{inputenc}

\usepackage{microtype}

\usepackage{inconsolata}
\usepackage{tikz}
\usepackage{comment}
\usepackage{booktabs}
\usepackage{amsmath,amssymb} % define this before the line numbering.
\usepackage{color}
\usepackage[capitalize]{cleveref}
\usepackage{algpseudocode}
\usepackage{multirow}
\usepackage{multicol}
\usepackage{makecell}
\usepackage{enumitem}
\usepackage{float}
\usepackage{subfigure}
\usepackage{wrapfig}
\usepackage{bbding}
\usepackage[normalem]{ulem}
\usepackage{balance}
\usepackage{adjustbox}

\title{Movie101: A New Movie Understanding Benchmark}

\author{Zihao Yue, Qi Zhang, Anwen Hu, Liang Zhang, Ziheng Wang, Qin Jin\thanks{*Corresponding Author.}\\
School of Information, Renmin University of China\\
\texttt{\{yzihao, zhangqi1996, anwenhu, zhangliang00, zihengwang, qjin\}@ruc.edu.cn}}

\begin{document}
\maketitle
\begin{abstract}
To help the visually impaired enjoy movies, automatic movie narrating systems are expected to narrate accurate, coherent, and role-aware plots when there are no speaking lines of actors.
Existing works benchmark this challenge as a normal video captioning task via some simplifications, such as removing role names and evaluating narrations with ngram-based metrics, which makes it difficult for automatic systems to meet the needs of real application scenarios. To narrow this gap, we construct a large-scale  Chinese movie benchmark, named \textbf{Movie101}. 
Closer to real scenarios, the Movie Clip Narrating (MCN) task in our benchmark asks models to generate role-aware narration paragraphs for complete movie clips where no actors are speaking. External knowledge, such as role information and movie genres, is also provided for better movie understanding. Besides, we propose a new metric called Movie Narration Score (MNScore) for movie narrating evaluation, which achieves the best correlation with human evaluation.
Our benchmark also supports the Temporal Narration Grounding (TNG) task to investigate clip localization given text descriptions. For both two tasks, our proposed methods well leverage external knowledge and outperform carefully designed baselines. 
The dataset and codes are released at \url{https://github.com/yuezih/Movie101}. 
\end{abstract}

\section{Introduction}

\begin{figure*}[t]
	\begin{center}
		\includegraphics[width=1\linewidth]{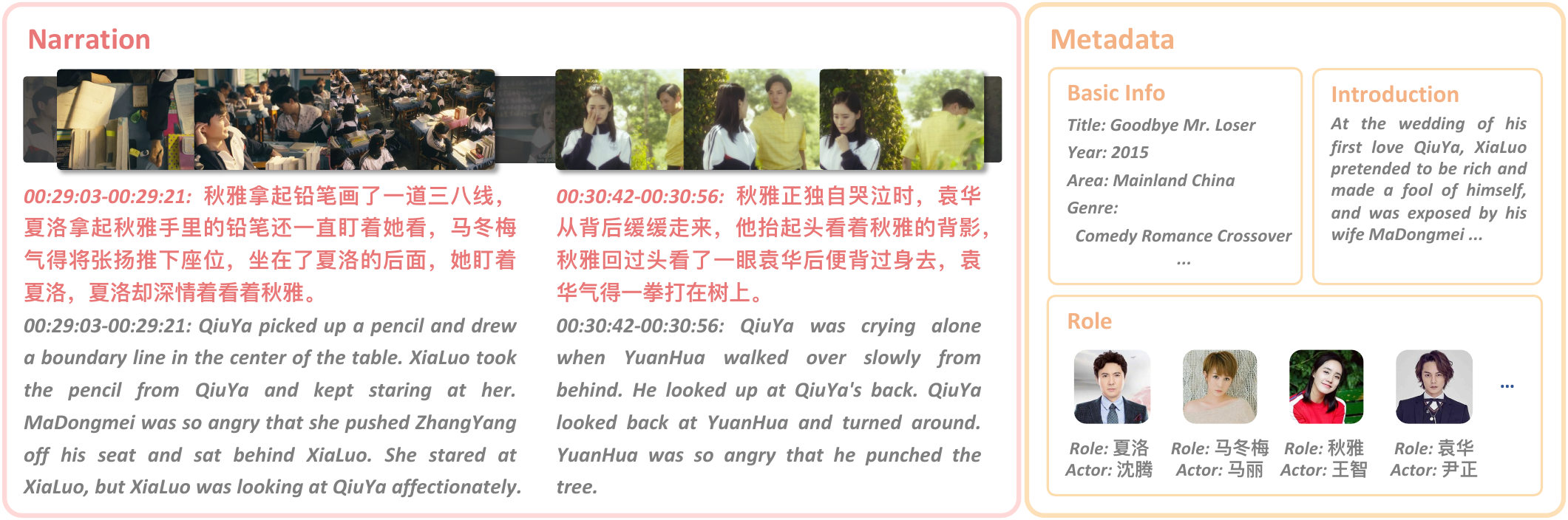}
	\end{center}
	\vspace{-8pt}
	\caption{Data samples from the movie \textit{Goodbye Mr. Loser}. (English translations are provided for easy reading.) }
	\label{fig:1}
\end{figure*}

The estimated number of visually impaired people worldwide was about 285 million by 2020, according to reports~\cite{vision_report}. While regulations are in place to ensure increased access for these audiences to experience the culturally dominant movies and TV shows on popular media platforms, technologies that provide them with genuine experience are becoming increasingly important. Audio description (AD, also known as video description) is a form of such technology intended for visually impaired audiences to experience the movie or TV show by hearing what is happening on-screen. However, producing movie narration scripts is not trivial, often requiring a professional writer to oversee the original movie. The high cost of narration generation \cite{Lakritz2006SG} greatly hinders the production of movies with AD and thus limits the opportunities for visually impaired users to experience movies. 

To address this issue, attempts have been carried out to automate AD production. Datasets of movies with ADs are constructed to support the research on automatic AD generation, including the MPII-MD dataset \cite{mpii-md_dataset} and M-VAD dataset \cite{m-vad_dataset}, with shot-level ADs or scripts aligned to the visual contents of movie. Consequently, different solutions for automatic movie narrating have been proposed based on these datasets~\cite{rohrbach_movie_2017}. 

However, existing benchmarks suffer from several limitations. Firstly, there is a gap between the designed tasks and the actual movie narration scenario. These tasks mainly focus on generating single-sentence narrations for shots of a few seconds. They can not support the generation of coherent narrations for longer plots, which is critical for the visually impaired to better understand the movie, and the timestamps of these shots are carefully annotated, which are difficult to obtain for new movies in real application. Meanwhile, these tasks treat the very distinctive movie narrating task as a normal video captioning task through some simplifications such as replacing role names with \texttt{SOMEONE}, resulting in the inability to connect roles to plots. Secondly, these benchmarks evaluate the generated narrations with ngram-based metrics, which can over-penalize a semantically correct but textually inconsistent narration, especially when there is only one reference available. 
In addition, these existing datasets are all in English. However, about one-fifth of the world's population speaks Chinese as their mother tongue, of whom more than 17 million are visually impaired \cite{blind_data}. 
Therefore, building a Chinese movie narration benchmark is necessary.

Intending to address the limitations of the existing narrating benchmarks, in this work, we propose a new benchmark with 101 Chinese movies for movie understanding, named \textbf{Movie101}. We collect the movies from the barrier-free channel on Xigua Video platform\footnote{\url{https://www.ixigua.com/channel/barrier_free}}, where normal movies are remastered with ADs. Through automatic process and manual correction, we obtain the ADs and actor lines from the raw videos. We crawl rich meta-information relevant to the movies as well. Finally, Movie101 contains 30,174 narration clips totaling 92 hours, with data samples as shown in \cref{fig:1}. 
As our investigation shows that narrations mostly occur at those times when no actors are speaking (see \cref{narration_distribution}), to achieve realistic movie narrating, we propose the Movie Clip Narrating (MCN) task that requires a model to narrate where there are no lines. It brings a potential benefit for identifying where to narrate in an unlabeled new movie, since the timestamps of the actor lines are easily accessible\footnote{The timestamps of the lines can be obtained from the movie script or by automatic methods such as OCR and ASR.}. Meanwhile, in order for the audience to accurately comprehend the role-related plots, concrete role names should be contained in the generated narration. For the MCN task, we reorganize the Movie101 dataset, merging the narration clips between two actor dialogues into a longer clip, to simulate real-scenario movie narrating. We thus obtain 14,109 long clips of variable length for narration generation. 
Moreover, to better evaluate the quality of model-generated narrations, we conduct human evaluations and design a new metric specific to movie narrating, namely Movie Narration Score (MNScore), which well aligns with human evaluation. 
In addition to the MCN task, our dataset also supports the Temporal Narration Grounding (TNG) task, which asks a model to locate target clips in the movie according to some text descriptions. 
For both tasks, we benchmark the performance of existing methods, and further propose our improved models by incorporating auxiliary external knowledge. 
In addition to MCN and TNG tasks, Movie101 can also potentially support other movie understanding tasks such as visual question answering and action recognition, etc.

The main contributions of this paper are as follows: 
1) We propose a new benchmark for movie understanding, Movie101, with a large number of video-aligned text descriptions in Chinese. 
2) We propose two primary tasks, MCN and TNG, and a new narrating evaluation metric MNScore, where MCN is more in line with the needs of actual movie narrating, while MNScore is more consistent with human evaluation. 
3) We benchmark state-of-the-art models and propose improved models enhanced by external knowledge for MCN and TNG, respectively. 
We expect our proposed Movie101 benchmark can inspire more explorations on narrating and understanding a whole movie. 
\section{Related Works}

\noindent \textbf{Datasets.}
Existing datasets to support the automatic narration generation task include M-VAD~\cite{m-vad_dataset} and MPII-MD~\cite{mpii-md_dataset}, which are merged into LSMDC \cite{rohrbach_movie_2017}. M-VAD, which is collected based on an automatic AD segmentation and alignment method, contains 47K videos from 92 DVDs, with an average length of 6.2s, each with an aligned narration. MPII-MD contains 68K videos from 94 movies with an average duration of 3.9s, about half of which come with paired scripts and the other half with paired ADs. In addition to movies, TV shows are also good data sources for automatic narration generation. \citet{lei2020tvr} propose TV Show Caption (TVC), a variant of TV Show Retrieval (TVR). It contains 11K short videos averaging 9.1s in length, and 26K captions describing the visual content, dialogues, and subtitles. All the existing datasets are in English.

\noindent \textbf{Video Captioning.}
As a classic vision and language task, the video captioning task requires a model to generate natural language descriptions for given videos. Solutions for normal video captioning go through stages from pre-designed templates \cite{kojima_2002, youtube2text} to sequence-to-sequence generation with deep neural networks \cite{pasunuru_2017}. 
A challenging variant for this task is dense video captioning \cite{anet_caption}, which requires the generation of multi-sentence descriptions for long multi-event videos. The two-stage generation approach, which firstly performs proposal detection on the video and then generates descriptions for each proposal separately, has been the dominant approach \cite{anet_caption, dense_park_2018, dense_senina_2014, dense_xiong_2018}. 
Recently, some works avoid event detection and generate paragraph descriptions directly based on the video, such as the one-stage paragraphing model (OVP) \cite{video_paragraph}, obtaining competitive performance compared to previous works, inspired by which we propose our knowledge-enhanced movie narrating model. 
Identity-aware video description that distinguishes different persons is more practical in real applications. \citet{park2020identity} attempt to achieve role-aware movie narrating by distinguishing different people using labels such as \texttt{PERSON1}, \texttt{PERSON2}, etc. However, it fails to generate concrete role names and falls short in terms of practicality.

\noindent \textbf{Temporal Sentence Grounding.}
The temporal sentence grounding (TSG) task aims to localize the moment in a video based on a natural language query~\cite{tsg_gao_2017}. A two-step pipeline has been the mainstream approach, which first produces a large number of moment candidates via sliding windows, then ranks them with their similarity to the query sentence. The following works try to improve the grounding performance by enhancing interaction between video and query modalities \cite{liu2021_ianet,Li_2022_tsg_visa} or introducing novel detection heads \cite{lei2021_mdetr,zhang2020_2dtan}. Specifically, for interaction methods, \citet{liu2021_ianet} adopt an Iterative Alignment Network (IA-Net) to iteratively interact inter- and intra-modal features within multiple steps. \citet{Li_2022_tsg_visa} explicitly decompose video and query into multiple structured hierarchies and learn fine-grained semantic alignment among them. 
In this work, we propose to incorporate external knowledge based on the IA-Net model structure.

\section{Dataset}

\begin{table*}[htb]
\caption{\label{tab:Movie101_stat}
Movie101 and other Movie Narrating and Temporal Sentence Grounding datasets. (* indicates statistics based on Chinese characters.)}
\vspace{-8pt}
\centering
\small
\begin{adjustbox}{width=\textwidth}
\begin{tabular}{@{}c|c|cccccc@{}}
\toprule
Task & Dataset & Video num. & Text num. & Avg. video len. & Avg. text len. & Avg. actions & Avg. role names \\ \midrule
\multirow{4}{*}{Narrating} & M-VAD & 47K & 47K & 6.2  sec. & 10.8 & - & - \\
 & MPII-MD & 68K & 68K & 3.9  sec. & 9.6 & 1.4 & 0.37 \\
 & TVC & 109K & 262K & 9.1  sec. & 13.4 & 1.9 & 0.75 \\ \cmidrule(l){2-8} 
 & \textbf{Movie101-N} & 14K & 14K & 20.4 sec. & 80.7* & 12.3 & 2.0 \\ \midrule
\multirow{4}{*}{Grounding} & Charades-STA & 10K & 16K & 31  sec. & 7.2 & 1.1 & 0 \\
 & ActivityNet & 20K & 72K & 118  sec. & 13.5 & 2.1 & 0.02 \\
 & TVR & 22K & 109K & 76  sec. & 13.4 & 1.9 & 0.75 \\ \cmidrule(l){2-8} 
 & \textbf{Movie101-G} & 101 & 30K & 6,144 sec. & 47.3* & 6.9 & 1.1 \\ \bottomrule
\end{tabular}
\end{adjustbox}
\end{table*}

\subsection{Data Collection}

\noindent \textbf{Movie Acquisition. }
To the best of our knowledge, there are only a handful of platforms that provide accessible movies in Chinese. 
The barrier-free channel of Xigua Video is one such platform that provides over 100 accessible movies online, and new movies are still being released that can support further expansion of our dataset. 
From Xigua Video, we collect all 101 movies available to date and crawl as much meta information as possible for each movie, including title, introduction, genres, directors, actors, etc. 
We emphasize actors in particular, including actor names, role names, actor portraits, role rankings, and other information about important roles. We expect such information can benefit the movie narrating task and general movie understanding tasks. 

\noindent \textbf{Narrations and Lines Extraction. }
As the movie lines and narrations are only available in the subtitle and audio format respectively from the platform, we therefore leverage OCR and automatic speech recognition (ASR) tools for transcription. For lines, we extract text from subtitles by open-source OCR toolkit PaddleOCR\footnote{\url{https://github.com/PaddlePaddle/PaddleOCR}} at 2.4 FPS, and manually remove the irrelevant subtitles from the beginning and the end of each movie. For narrations, we extract the audio track from the movie and utilize the ASR service provided by iFlyTek\footnote{\url{https://www.xfyun.cn/doc/asr/lfasr/API.html}}, which detects the speech in the audio and transcribes it into text. In addition, the service supports identifying different speakers, which helps discriminate the narrator from the actors. However, the ASR service is not perfect, and its outputs contain errors such as wrong characters, unreasonable sentence breaking, and misidentification of narrations as movie dialogues, etc. Therefore, we recruit human annotators to further correct the ASR transcription errors and remove non-narration texts manually to improve the data quality. We also delete the irrelevant fragments at the beginning (e.g., movie synopsis, cast introductions) and the summary narration at the end. For coherency, we further organize the narration fragments at the clip level. We merge every two fragments if their temporal gap is less than 1 second. we also apply a paragraph-length threshold of 100 characters to limit over-merging to avoid excessively long clips. We take punctuation into account as well, for example, a period in Chinese is likely to mean the end of a narrative paragraph. Further detailed descriptions of data quality can be found in \cref{data_quality}. 

\begin{figure}[t]
	\begin{center}
		\includegraphics[width=1\linewidth]{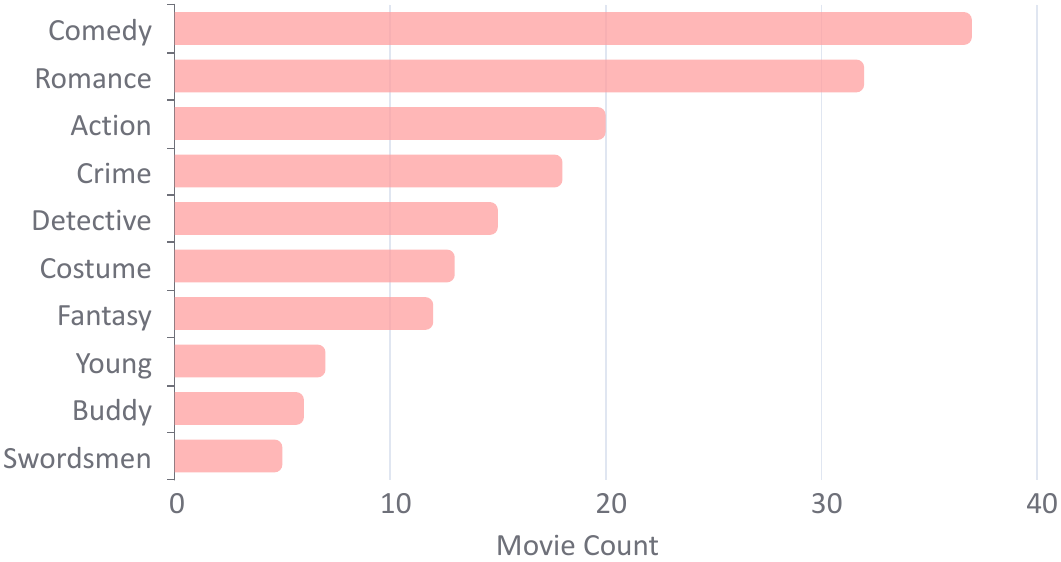}
	\end{center}
	\vspace{-8pt}
	\caption{Distribution of movie genres. }
	\label{fig:genres_stat}
\end{figure}

\begin{figure}[t]
	\begin{center}
		\includegraphics[width=0.95\linewidth]{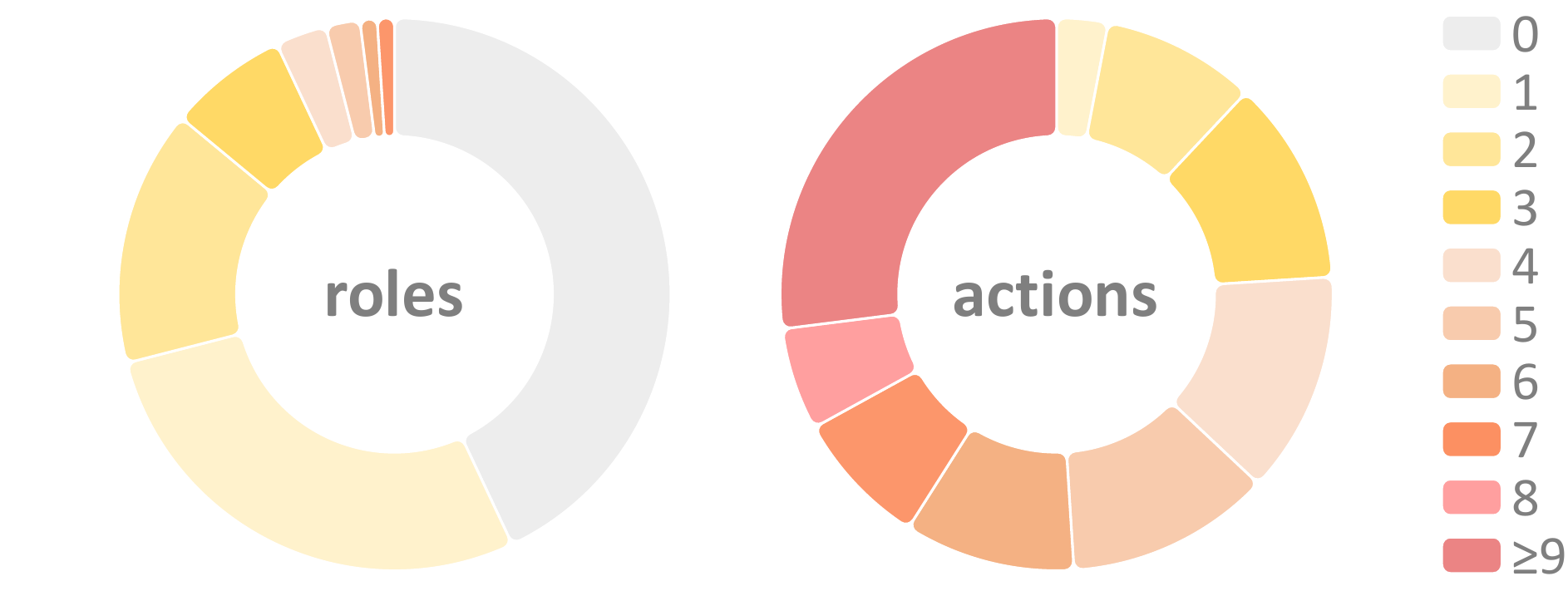}
	\end{center}
	\vspace{-8pt}
%	\caption{Number distribution of role names and actions in each narration in Movie101.}
\caption{Distribution of the number of role names and the number of actions in each narration in Movie101.}
	\label{fig:role_verb_stat}
\end{figure}

\begin{figure}[t]
	\begin{center}
		\includegraphics[width=1\linewidth]{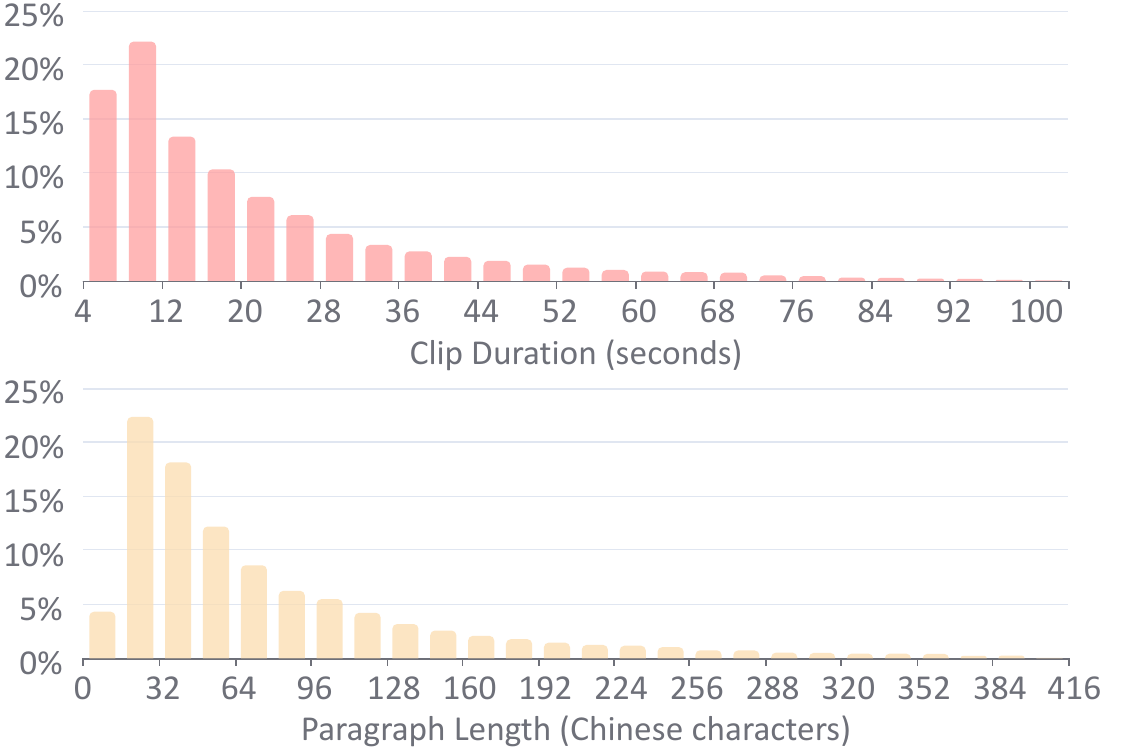}
	\end{center}
	\vspace{-8pt}
	\caption{Length and duration distribution of narration clips in Movie101-N. }
	\label{fig:len_stat}
\end{figure}

\noindent \textbf{Movie101-N and Movie101-G. }
For real-life movie narrating, models are expected to narrate in the breaks between different actor dialogues. Thus, we reorganize Movie101 to fit this task format. Concretely, we first merge the independent lines in Movie101 into dialogues, where two lines with a temporal gap shorter than 5 seconds are considered to belong to one dialogue. Then, we merge all the narration clips between two adjacent dialogues into a long paragraph. In this way, we obtain \textbf{Movie101-N} with narration paragraphs separated by dialogues, which well simulates the practical narrating challenge. Meanwhile, with rich video-text pairs in Movie101, we create another variant dataset to support the temporal grounding tasks, named \textbf{Movie101-G}, where narrations are taken as queries and aligned videos serve as targets. For validation and testing, we carefully select 10 movies of different genres for each respectively. 

\begin{figure*}[t]
	\begin{center}
		\includegraphics[width=1\linewidth]{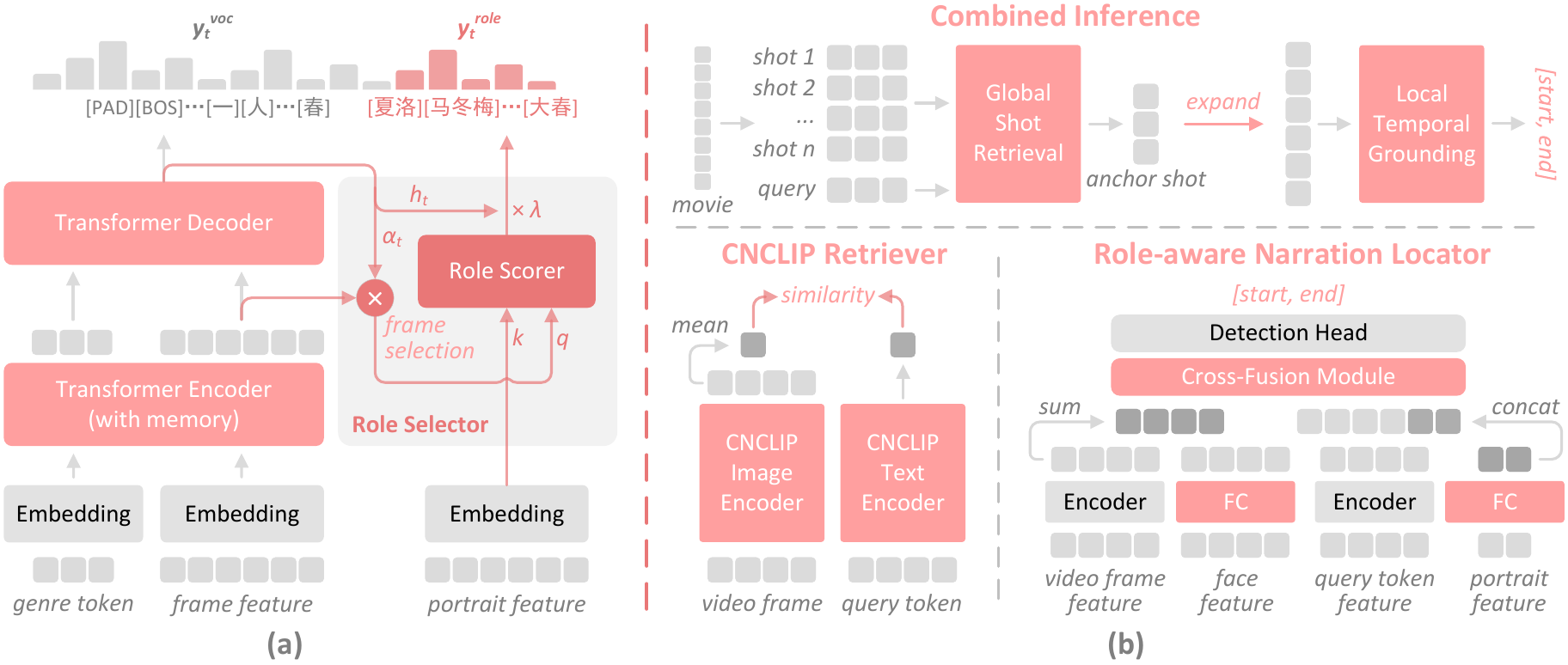}
	\end{center}
	\vspace{-8pt}
	\caption{The frameworks of our models. (a) Role-pointed Movie Narrator (RMN) for the Movie Clip Narrating task, and (b) Global Shot Retrieval + Local Temporal Grounding for the Temporal Narration Grounding task. }
	\label{fig:model}
\end{figure*}

\subsection{Dataset Statistics}

\noindent \textbf{Movie Properties.}
Movie101 contains 101 movies, involving 41 genres (a movie can belong to up to 4 genres) and 645 roles in total. \cref{fig:genres_stat} shows the numbers of movies in the top 10 most popular genres, with comedy, romance, and action in the top 3. %\cref{fig:genres_stat} (b) shows the distribution of the number of roles in different movies, with more than 52.5\% of movies containing more than 5 roles. 

\noindent \textbf{Clip Properties.}
Movie101 contains a total of 30,174 short narrated clips with an average duration of 11.0 seconds and an average length of 47.3 Chinese characters. 
From the narrations, we locate role names and action words with the metadata and the Chinese Part-Of-Speech (POS) tagging toolbox HanLP\footnote{\url{https://github.com/hankcs/HanLP}}, to detail role and action content in narrations. \cref{fig:role_verb_stat} shows the distribution of the number of role names and actions per single clip. 
The narrating variant dataset Movie101-N contains 14,109 long narration clips of an average length of 20.4 seconds and 80.7 characters. The comparison in \cref{tab:Movie101_stat} shows that Movie101-N contains much longer video clips and text descriptions than existing movie narrating datasets, while the length distribution in \cref{fig:len_stat} indicates that the clip length varies a lot. Movie101-G contains 30,174 clips to be located from 101 movies. The average video length of 6,144 seconds also greatly exceeds existing TSG datasets.
\section{Movie Clip Narrating}

\subsection{Task Description}
In order to help the visually impaired keep up with the plot in the movie, we first propose a Movie Clip Narrating (MCN) task, which aims to generate a plot-related paragraph description given a clip in Movie101-N.
Besides, the narration styles may vary across different genres of movies. The role portraits are important external knowledge for a model to accurately describe the subject of actions. Thus, we also provide this information in Movie101-N to support the MCN task.

\subsection{Proposed Method}
For the MCN task, with multimodal inputs including video, movie genres, role names, and actor portraits, we propose a Transformer-based \cite{vaswani2017attention} model with an encoder-decoder framework, namely Role-pointed Movie Narrator (RMN), where the encoder mainly encodes video clips and the decoder generates narrations, as shown in \cref{fig:model} (a). 

On the encoder side, taking into account the frame-level visual information, the video clip is embedded into a sequence of frame-level features. To emphasize the roles, we extract face features from each frame and concatenate them to the corresponding frame feature sequentially based on the confidence scores of face detection. With learnable genre embeddings, genres are also represented as a sequence of genre features. After video and genre representation, we apply a Transformer encoder to perform cross-encoding. Then, we follow the One-stage Video Paragraphing model (OVP) \cite{video_paragraph} to use a dynamic memory bank to refine the video-part representations, which updates at each decoding step.

On the decoder side, in addition to the Transformer decoder, we enable the model to directly choose a complete role name from the movie cast according to context during token-by-token generation via a pointer network \cite{ptr_gu_2016}. At the decoding step $t$, with the decoder hidden state $h_{t}$, we first calculate the token scores $y^{voc}_{t}$ among normal vocabulary. Then we design a Role Selector module to get the name scores among external role vocabulary. Concretely, with the decoder's video-part attention distribution $\alpha_{t}$, we perform a weighted summation among video representations to get a context-filtered video feature. Then the role scores $y^{role}_{t}$ are computed with the context-filtered video feature as query and portrait features as key. Finally, the prediction distribution at step $t$ is calculated as follows:
\begin{gather}
y_{t}=f([y^{voc}_{t};\lambda y^{role}_{t}]) 
\end{gather}
where $[;]$ means concatenation, $\lambda$ is a gate computed from $h_{t}$, $f()$ is the softmax function. 

\begin{table}[t]
\caption{\label{tab:metric_acc}
%Accuracy of metrics to evaluate the candidate narrations. (Acc.: accuracy; Info.: informativeness; Qual.: textual quality)
Accuracy of metrics in terms of their assessment of the candidate narrations against human assessment. (Acc.: accuracy; Info.: informativeness; Qual.: textual quality)
}
\vspace{-8pt}
\centering
\small
\begin{tabular}{@{}l|ccc|c@{}}
\toprule
Metric & Acc. & Info. & Qual. & Overall \\ \midrule
CIDEr & 86.7 & 83.0 & 82.0 & 87.0 \\
BLEU@4 & 85.0 & 82.0 & 80.3 & 86.0 \\
METEOR & 87.0 & 82.7 & 82.7 & 87.0 \\ \midrule
CLIPScore & 39.7 & 40.0 & 38.0 & 39.0 \\
BERTScore & 88.0 & 84.7 & \textbf{87.7} & 90.3 \\
EMScore & 40.3 & 42.3 & 41.3 & 41.3 \\
DIV & 51.7 & 51.3 & 57.3 & 54.7 \\
PPL & 43.0 & 46.7 & 45.0 & 45.3 \\
RoleF1 & 33.0 & 31.0 & 28.3 & 32.3 \\ \midrule
\textbf{MNScore} & \textbf{90.3} & \textbf{86.7} & 86.3 & \textbf{92.0} \\ \bottomrule
\end{tabular}
\end{table}

\begin{table*}[t]
\caption{\label{tab:mcn_main}
Movie Clip Narrating Performance on Movie101-N. ($f_v$: face features from the video; $g$: movie genres)
}
\vspace{-8pt}
\centering
\small
\begin{tabular}{@{}c|cc|ccc|c@{}}
\toprule
Model & $f_v$ & $g$ & EMScore & BERTScore & RoleF1 & MNScore \\ \midrule
Vanilla Transformer &  &  & 0.153 & 0.150 & 0 & 12.55 \\
OVP &  &  & \textbf{0.155} & 0.159 & 0 & 13.18 \\ \midrule
\multirow{3}{*}{RMN} &  &  & 0.153 & 0.185 & 0.195 & 18.13 \\
 & \checkmark &  & 0.154 & 0.186 & \textbf{0.240} & 18.97 \\
 & \checkmark & \checkmark & 0.154 & \textbf{0.188} & 0.238 & \textbf{19.07} \\ \bottomrule
\end{tabular}
\end{table*}

\subsection{Evaluation}
\label{mcn_eval}

Existing movie narration benchmarks directly adopt ngram-based metrics including CIDEr, BLEU, and METEOR as in normal video captioning. However, there are pitfalls for these metrics, such as underestimating semantically correct but textually inconsistent phrases, which have been widely reported \cite{zhang2019bertscore, shi2022emscore}. For movie narrating, a movie clip can be narrated in multiple expressions, while there is only one reference. Thus, text matching is inadequate to measure the quality of a narration paragraph.

To better evaluate the generated narrations in the MCN task, we conduct a manual evaluation to investigate how humans assess different narrations. We randomly select 30 movie clips, each with 5 candidate narrations, of which 3 are derived from the predictions of different models and 2 are obtained by disturbing the ground truth narrations. %ground truth deprivation. 
Next, we recruit 10 annotators to individually rank the candidates for each video in terms of accuracy, informativeness, and textual quality. Accuracy defines how the narration accurately describes the video, especially roles, actions, and objects; informativeness defines how richly the narration reveals the video content; textual quality is determined by the narration fluency and grammatical correctness. 

With the human evaluation results, we investigate a wide range of objective metrics as follows: (1) State-of-the-art video captioning metrics based on deep neural networks including CLIPScore \cite{hessel2021clipscore}, BERTScore \cite{zhang2019bertscore} and EMScore \cite{shi2022emscore}, which are reported outperforming ngram-based metrics in video captioning evaluation; (2) Textual quality metrics including n-grams diversity\cite{div} (DIV) and causal language model perplexity (PPL); (3) F1 score of role name generation (RoleF1). For every two candidate narrations of a video, we use human ranking as a reference to determine whether these metrics correctly judge which of the two candidates is better or worse, and the accuracy is used for evaluating metrics' correlation with human judgment. Finally, we settle on a new metric Movie Narration Score (MNScore) as follows: 

\begin{gather}
mns=\frac{1 \cdot ems + 4 \cdot berts + 1 \cdot rf1}{6}\times 100
\end{gather}
where $mns$, $ems$, $berts$ and $rf1$ refer to MNScore, EMScore, BERTScore and RoleF1, respectively. As shown in \cref{tab:metric_acc}, BERTScore outperforms ngram-based metrics in narration evaluation accuracy, while our new proposed MNScore achieves the best alignment with human evaluation. 
More details about the implementation of the candidate narrations and the above metrics are presented in \cref{imple_detail}. 

\subsection{Experiments}
\label{mcn_experiments}

\noindent \textbf{Implementation Details.}
In our proposed method, models are trained with next-token language modeling by the maximum likelihood estimation (MLE) objective. For videos, we use CLIP \cite{clip} pre-trained on large-scale image-text pairs and MIL-NCE \cite{mil-nce} pre-trained on HowTo100M videos \cite{howto100m} to extract frame-level CLIP and S3D features with dimensions of 512 and 1024, respectively, at 1 FPS, and further concatenate them. For faces in video frames and portraits, we use the Arcface model \cite{arcface} pre-trained on MS1M \cite{ms1m} to extract face features. When there are insufficient faces detected within a frame, the face feature extractor compensates by substituting the extracted features with zero-vectors. %All models are trained with Movie101 as augmentation data beyond Movie101-N. %More details can be found in \cref{supp_inple_details_mcn}.

\begin{table}[t]
\caption{\label{tab:retrieval}
Global Shot Retrieval performance of the first-stage model on Movie101-GSR(temp). %\texttt{black}: performance averaged by movie; \textcolor[RGB]{84,130,53}{\texttt{green}} / \textcolor[RGB]{192,0,0}{\texttt{red}}: highest / lowest single movie performance surpluses / reduction.  
}
\vspace{-8pt}
\centering
\small
\begin{tabular}{@{}c|ccc@{}}
\toprule
Model & Recall@1 & Recall@5 & Recall@10 \\ \midrule
CNCLIP & 25.98 & 54.91 & 66.99 \\ \bottomrule
\end{tabular}
\end{table}

\noindent \textbf{Results \& Analysis.} We choose Vanilla Transformer \cite{mtrans_zhou_2018} and state-of-the-art video paragraphing model OVP~\cite{video_paragraph} as the MCN baselines. 

As shown in \cref{tab:mcn_main}, RMN outperforms the baselines by a large margin, especially on RoleF1. This indicates that our model learns to generate role names from external knowledge with the help of the pointer network. To verify the contribution of the genre and face representations in our RMN model, we also perform an ablation study by progressively adding these representations as input. 
From the results, face features extracted from video frames bring significant gains in role awareness, which shows that using face features to bridge the video content and external actor portraits is beneficial for generating role-related narrations. Qualitative results can be found in \cref{qualitative}.

\section{Temporal Narration Grounding}

\subsection{Task Description}
To help people locate clips of interest during movie entertainment, an AI agent should be able to understand users' intentions and locate the target clips. To achieve this goal, we propose the Temporal Narration Grounding (TNG) task. Given a clip narration as the query, TNG aims to predict the starting and ending time of the clip in the whole movie.

\subsection{Proposed method}
\label{tng_method}

Existing temporal sentence grounding models can hardly handle an entire movie input with limited computational resources. Thus, we propose a two-stage framework for the TNG task, with global shot retrieval to coarsely locate the target clip in the first stage and local temporal grounding to finalize the precise timestamp of the target clip in the second stage, as shown in \cref{fig:model} (b). 

\noindent \textbf{Global Shot Retrieval. } To find the approximate location of the target, we treat it as a text-video retrieval subtask. We divide a movie into 20s-long shots, and the shot with the highest similarity to the text query will be used as the anchor for further grounding in the second stage. For training such a retrieval system, we construct a temporary dataset Movie101-GSR(temp). Concretely, after cutting the movie into shots, each shot and each annotated narration in Movie101 are judged with the temporal overlap whether they can be considered as an aligned video-text pair.\footnote{A shot and a narration with a temporal overlap larger than half of the duration of either the shot or the narration are regarded as aligned. } 

\begin{table}[t]
\caption{\label{tab:grounding}
Local Temporal Grounding performance of the second-stage models on Movie101-LTG(temp). ($f_v$ and $f_t$ refer to adding face features to the video and text representations, respectively.)
}
\vspace{-8pt}
\centering
\small
\begin{tabular}{@{}c|cc|cc|cc@{}}
\toprule
\multirow{2}{*}{Model} & \multirow{2}{*}{$f_v$} & \multirow{2}{*}{$f_t$} & \multicolumn{2}{c|}{Rank@1} & \multicolumn{2}{c}{Rank@5} \\ \cmidrule(l){4-7} 
 &  &  & IoU0.3 & IoU0.5 & IoU0.3 & IoU0.5 \\ \midrule
2D-TAN &  &  & 25.85 & 18.60 & 52.17 & 43.82 \\
IA-NET &  &  & 25.16 & 17.98 & 57.11 & 42.68 \\ \midrule
RNL & \checkmark &  & 26.64 & 19.01 & \textbf{59.63} & 44.51 \\
RNL &  & \checkmark & 16.98 & 19.57 & 57.18 & 42.86 \\
RNL & \checkmark & \checkmark & \textbf{27.54} & \textbf{20.22} & 59.52 & \textbf{45.69} \\ \bottomrule
\end{tabular}
\end{table}

\begin{table*}[htb]
\caption{\label{tab:tng_combine}
Combined inference performance of our proposed two-stage method on Movie101-G.
}
\vspace{-8pt}
\centering
\small
\begin{tabular}{@{}c|c|cccc|cccc@{}}
\toprule
\multirow{2}{*}{Model} & \multirow{2}{*}{\begin{tabular}[c]{@{}c@{}}$k$-way\\ re-ranking\end{tabular}} & \multicolumn{4}{c|}{Rank@1} & \multicolumn{4}{c}{Rank@5} \\ \cmidrule(l){3-10} 
 &  & IoU0.1 & IoU0.3 & IoU0.5 & IoU0.7 & IoU0.1 & IoU0.3 & IoU0.5 & IoU0.7 \\ \midrule
\multirow{3}{*}{CNCLIP+RNL} & 1 & \textbf{18.69} & \textbf{11.65} & \textbf{6.66} & \textbf{15.38} & 35.79 & 29.77 & 22.68 & \textbf{14.87} \\
 & 2 & 18.17 & 10.53 & 5.99 & 14.45 & 36.98 & \textbf{30.98} & \textbf{26.28} & 13.56 \\
 & 3 & 17.18 & 10.05 & 5.47 & 13.96 & \textbf{37.91} & 30.23 & 25.37 & 13.33 \\ \bottomrule
\end{tabular}
\end{table*}

We build the retrieval model by transferring a Chinese Vision-Language Pre-training (VLP) model ChineseCLIP \cite{yang2022chineseclip} (CNCLIP) from image-text to video-text. Specifically, the shot frames are separately encoded as image features by the visual encoder of CNCLIP, and the final video feature is obtained by performing mean pooling over the \verb|CLS| tokens of all frames. We then perform contrastive learning between the video and text features on Movie101-GSR(temp) to fine-tune the modified CNCLIP.

\noindent \textbf{Local Temporal Grounding. }
After obtaining the anchor shot in the first stage, we further localize the target clip within a 200-second window around the anchor shot. This requires the temporal sentence grounding in a 200s-long movie clip, where comprehending the actions of different roles is critical. Therefore, based on the state-of-the-art TSG model IA-Net \cite{liu2021_ianet}, we propose Role-aware Narration Locator (RNL). 
% IA-Net consists of three components: (1) visual and textual encoder, which encode video frame features and query token features respectively; (2) cross-fusion module, which iteratively interacts inter- and intra-modalities in multiple steps for semantic alignment; (3) detection head, which predicts the temporal offsets against anchors with respective confidence scores. 
% While IA-Net encodes the input frame features to final frame representation $V$ before cross-modal fusion, 
With a bi-directional GRU \cite{chung2014empirical} visual encoder, we encode the input frame features to get temporal context-aware frame representations $V$. We in addition extract face features from the frames and encode them with a fully connected (FC) layer to filter key face information $F$. Then we finalize the visual representation by summing $V$ and $F$. For text encoding, to relate role names in the text query with roles in the video, we extract face features from the portraits that correspond to the role names and encode them as visual token representations with a FC layer, which are then concatenated to the query's textual token representation sequence. %In each training epoch, we randomly construct a 200s-long untrimmed movie clip covering the target for each query, to simulate the Local Temporal Grounding subtask. 
During training, for each target, we randomly select a 200s-long clip window that covers the target in each training epoch. We also construct a temporary dataset Movie101-LTG(temp) with fixed window to separately evaluate the second-stage model performance.

\subsection{Experiments}

\noindent\textbf{Implementation Details.}
For Global Shot Retrieval, we use average Recall@$n$ ($n\in{1,5,10}$) to evaluate the retrieval performance on all movies. For Local Temporal Grounding, following previous works~\cite{zhang2020_2dtan}, we use ``R@$n$, IoU@$m$'' as metrics, which are defined as the percentage of at least one of top-$n$ proposals having a larger temporal IoU than $m$ with the ground truth. We fine-tune CNCLIP-huge on our Movie101-GSR(temp) for Global Shot Retrieval, and benchmark two code-released state-of-the-art temporal grounding models 2D-TAN \cite{zhang2020_2dtan} and IA-Net\cite{liu2021_ianet} on Movie101-LTG(temp) for Local Temporal Grounding. In our RNL model, the video frame, face, and text feature extractors are pre-trained MIL-NCE, Arcface (same as in the MCN task) and BERT-base-Chinese \cite{devlin2018bert}, respectively. %More details are provided in Appendix. 

\noindent \textbf{Results \& Analysis.}
\cref{tab:retrieval} and \cref{tab:grounding} show the performance of models on Global Shot Retrieval and Local Temporal Grounding, respectively. Our RNL outperforms baselines by introducing role-aware video and text encoding, indicating that distinguishing actions of different roles is critical for grounding movie narration. Furthermore, we perform an ablation study to verify the effectiveness of role-aware encoding. As shown in \cref{tab:grounding}, adding face features to either video or text representations outperforms our base method IA-Net. RNL with both role-aware video and text encoding achieves the best performance. \cref{tab:tng_combine} shows the performance of combined inference by Global Shot Retrieval and Local Temporal Grounding. We in addition show the performance of $k$-way re-ranking, where the top-$k$ shots retrieved in the first stage are respectively used as the anchors in the second stage, and all predictions obtained are re-ranked with their confidence scores. The experimental results show that $k$-way re-ranking improves Rank@5 performance but harms Rank@1 performance. Qualitative results can be found in \cref{qualitative}.
\section{Conclusion}

In this work, we propose Movie101, a Chinese large-scale video benchmark for movie understanding. To assist visually impaired people in enjoying movies, we propose a more realistic Movie Clip Narrating task to address the automatic movie description issue and design a human-preference-compatible metric MNScore for narrating evaluation. Movie101 also supports the Temporal Narration Grounding task, which is more challenging than the previous TSG benchmarks. Furthermore, our experiments validate the importance of external knowledge including genres and roles for movie understanding. However, there is still a significant gap between our models and expert annotations. This reveals that further research endeavors are still needed to help visually impaired people enjoy movies by AI. 
\section*{Limitations}

Keeping narration coherent within a movie is crucial for visually impaired people to enjoy the movie. In this work, we move a step forward for this target by setting the ground-truth texts in the Movie Clip Narrating task as narration paragraphs and providing longer video clips as inputs.
However, how to ensure description coherence across different clips within a movie has not been studied in this work. This requires a higher-level comprehending ability of models to process the whole movie and connect different plots. We leave this to our future investigation.
\section*{Ethics Statement}
We propose Movie101, a new benchmark to support exploring technologies to benefit the accessibility of the visually impaired. There are two potential ethical issues with our work, regarding data source and crowdsourcing services, respectively. We state each of them as follows: 

\noindent \textbf{Data Source. } The collected movies are publicly available from Xigua Video, and are allowed to be crawled according to the service contract of the website\footnote{\url{https://www.ixigua.com/robots.txt}}. Considering the copyright issue, we will only release the url list of movies. Besides, our data source does not contain any information that names or uniquely identifiable individuals or offensive content. 

\noindent \textbf{Crowdsourcing Services. } We recruited 20 Chinese college students (12 females and 8 males) via social media. For ASR outputs cleaning, workers were required to correct errors in the narration text while watching the movie. For each movie, it took about 2 hours with a payment of 50 RMB (\$7.40 USD). To review corrections, for each movie, it took about 30 minutes with a payment of 25 RMB (\$3.70 USD). Our payment is fair and reasonable in China, especially since the work is easy and fun. Before the annotation works began, we introduced the future use of the data in the task document to ensure that everyone was informed.

\section*{Acknowledgements}
This work was partially supported by the National Key  R\&D  Program  of  China  (No.2020AAA0108600) and the National Natural Science Foundation of China (No. 62072462).

\bibliography{anthology,custom}
\bibliographystyle{acl_natbib}

% \clearpage
\appendix

\begin{table*}[t]
\caption{\label{tab:computation}
Key hyperparameters and computational burden for models training.
}
\vspace{-4pt}
\centering
\small
\begin{tabular}{@{}c|cccl@{}}
\toprule
Model & Batch size & Learning rate & Training epochs & \multicolumn{1}{c}{GPU hours / epoch} \\ \midrule
VT & $150$ & $1e\!-\!4$ & $\leq100$ & $\sim$3min on single RTX 2080ti \\
OVP & $56$ & $1e\!-\!4$ & $\leq100$ & $\sim$40min on single RTX 3090 \\
RMN & $56$ & $1e\!-\!4$ & $\leq100$ & $\sim$1h on single RTX 3090 \\
CNCLIP & $16$ & $2e\!-\!6$ & $\leq1$ & $\sim$1h on 4 RTX A6000 nodes \\
2D-TAN & $64$ & $1e\!-\!4$ & $\leq30$ & $\sim$40min on single RTX 3090 \\
IA-Net & $64$ & $8e\!-\!4$ & $\leq15$ & $\sim$20min on single RTX 3090 \\
RNL & $64$ & $8e\!-\!4$ & $\leq15$ & $\sim$20min on single RTX 3090 \\ \bottomrule
\end{tabular}
\end{table*}

\section{Narration Distribution}
\label{narration_distribution}
Clips where `no actors are speaking' refer to ANY scene wherein no verbal dialogue is being employed by the actors, regardless of whether they are visually present or absent. This definition encompasses, for example, a scene focused solely on a depiction of the sky. 
We detail the dialogues and narrations in the 101 collected movies. By merging the actor lines, we obtain a total of 15,307 dialogues, constituting 15,206 dialogue gaps with a total duration of 99.4 hours. The 30,174 narration clips we collect fill in 95.3\% of the dialogue gaps in terms of quantity and cover 92.9\% in terms of duration. Therefore, it is reasonable to assume that where there are no lines, there is a need for narration.

\section{Dataset Quality Description}
\label{data_quality}
We adopt a two-stage annotation process to ensure the quality of the narrations. In the first stage, a group of workers is recruited to clean the data according to our guidelines. In the second stage, another group of workers further checks and corrects the annotation data. Our heuristics used to divide the paragraphs are designed based on our observation experience. 
We further conduct a manual evaluation of the narration quality. Of the randomly sampled 300 paragraphs, (1) in terms of narration recognition, 96.7\% are textually consistent with original ADs; (2) as for the paragraph coherence, 90\% maintain complete and coherent semantics, 7.7\% should be merged with contexts, and 2.3\% should be divided into multiple paragraphs. Thus, the narration is of good quality to support downstream tasks. 

\section{Implementation Details}
\label{imple_detail}

\noindent \textbf{Candidate Narrations. } In \cref{mcn_eval}, We provide 5 different candidate narrations for each sampled movie clip for human evaluators to rank. These candidates are created as follows: 
\begin{enumerate}
\setlength{\itemsep}{0pt}
\setlength{\parsep}{0pt}
\setlength{\parskip}{0pt}
    \item generated by the Vanilla Transformer \cite{mtrans_zhou_2018}; 
    \item generated by the OVP model \cite{video_paragraph};
    \item generated by our proposed RMN model;
    \item generated by disturbing the ground truth with role name removal and replacement;
    \item generated by disturbing the ground truth with nouns and verbs replacement. 
\end{enumerate}

\noindent \textbf{Metrics Implementation. } For CLIP-based metrics including CLIPScore and EMScore, we fine-tune ChineseCLIP-huge \cite{yang2022chineseclip} on our dataset in the same way as in \cref{tng_method}. For each movie clip and generated narration, CLIPScore is calculated with the mean pooled feature of 10 uniformly selected frames and the overall text feature, while EMScore is calculated with all selected frame features and textual token features. For BERTScore, we use the BERT-base-Chinese \cite{devlin2018bert} model checkpoint to calculate, and rescale the raw BERTScore with baseline\footnote{\url{https://github.com/Tiiiger/bert_score/blob/master/journal/rescale_baseline.md}}. For DIV, we calculate 1-gram diversity and 2-gram diversity following \citet{div}, and average them. For PPL, we obtain the perplexity of each narration with the causal Ernie 3.0 model \cite{sun2021ernie} following the calculation of HuggingFace\footnote{\url{https://huggingface.co/spaces/evaluate-metric/perplexity}}. For RoleF1, we extract role names from the ground truth and the generated narration. We measure how the generated narration covers the roles appearing in the movie clip by Recall; given that these generated role names may also come from the model's hallucination, for example from a wrong movie, we also take Precision into account. Finally, we calculate the F1 score with Precision and Recall.

\noindent \textbf{Hyperparameters and Computation. } We detail the key hyperparameters and computational burden for the models training in \cref{tab:computation}. For each model, the results are derived from a single run. 

\section{Qualitative Result}
\label{qualitative}

\begin{figure*}[ht]
	\begin{center}
		\includegraphics[width=1\linewidth]{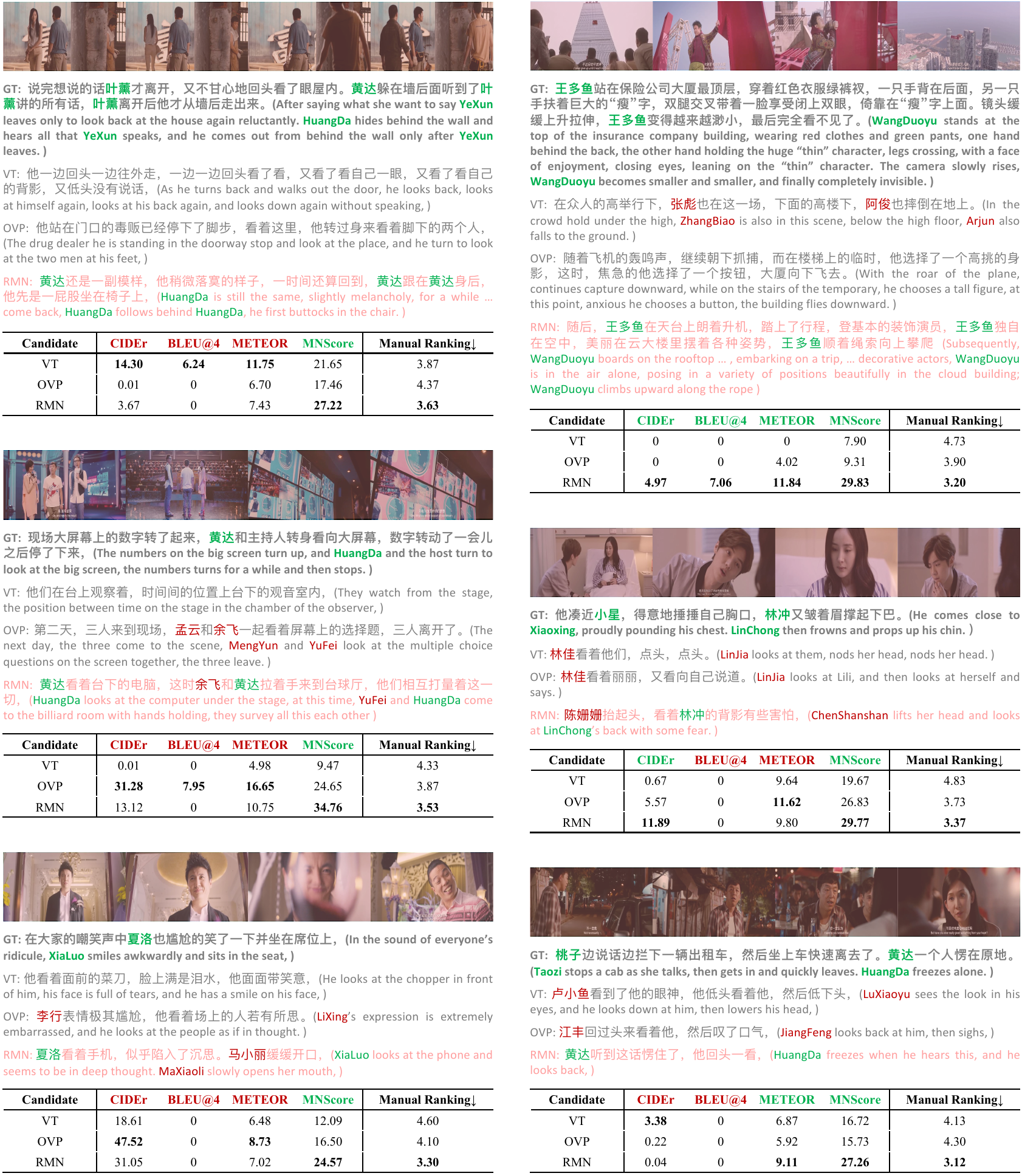}
	\end{center}
	\vspace{-8pt}
	\caption{Qualitative Results of the MCN task. (GT: the ground truth; VT: Vanilla Transformer). In the narration texts, \textcolor[RGB]{78,173,91}{green} and \textcolor[RGB]{176,36,24}{red} characters denote the correctly and wrongly generated role names, respectively. In the tables, metrics in \textcolor[RGB]{78,173,91}{green} indicate that the ranking of candidates by the metric is consistent with human ranking, while \textcolor[RGB]{176,36,24}{red} indicates inconsistency.}
	\label{fig:mcn_case}
\end{figure*}

\begin{figure*}[ht]
	\begin{center}
		\includegraphics[width=1\linewidth]{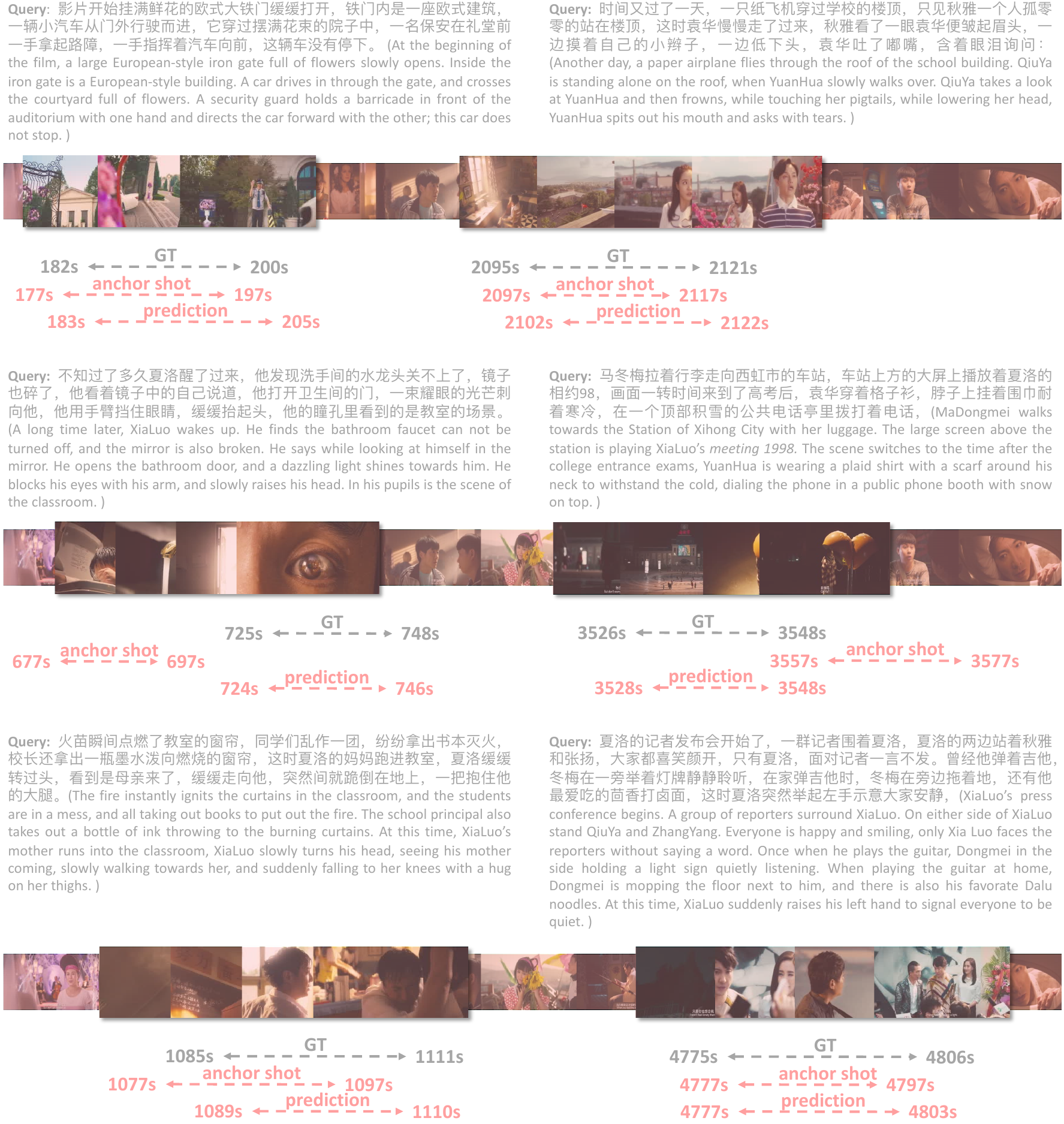}
	\end{center}
	\vspace{-8pt}
	\caption{Qualitative results of the TNG task from the movie \textit{Goodbye Mr. Loser}. }
	\label{fig:tng_case}
\end{figure*}

\subsection{Movie Clip Narrating}

\cref{fig:mcn_case} shows the qualitative results of the MCN task, including the generation results of baselines and our proposed RMN model, and the evaluation results of previous metrics and our proposed MNScore. Vanilla Transformer and OVP can correctly mention some actions but fail to generate correct role names because these roles never appear during training. However, with the help of the Role Selector module, our RMN could well relate roles in video clips with their role names. In addition, these cases demonstrate that our newly proposed MNScore evaluates more consistently with humans. 

\subsection{Temporal Narration Grounding}
\cref{fig:tng_case} shows the qualitative results of our proposed two-stage method. Through Global Shot Retrieval, we obtain an anchor shot near the target clip from the whole movie, which further helps Local Temporal Grounding to locate the final target.

\end{document}